\documentclass[conference]{IEEEtran}
\IEEEoverridecommandlockouts
\usepackage{cite}
\usepackage{amsmath,amssymb,amsfonts}
\usepackage{algorithmic}
\usepackage{graphicx}
\usepackage{textcomp}
\usepackage{xcolor}
\usepackage{booktabs}
\usepackage{array}
\usepackage[hidelinks]{hyperref}
\graphicspath{{figures/}}

\makeatletter
\newcommand{\linebreakand}{%
  \end{@IEEEauthorhalign}
  \hfill\mbox{}\par
  \mbox{}\hfill\begin{@IEEEauthorhalign}}
\makeatother

\newcommand{\NN}{Noise2Noise}
\newcommand{\NV}{Noise2Void}
\newcommand{\MAE}{\textsc{mae}}
\newcommand{\MSE}{\textsc{mse}}

\begin{document}

\title{Noise2Noise Revisited: Training Pair Distributions Dominate Loss Choice in Self-Supervised Denoising}

\author{%
\IEEEauthorblockN{Dingyan Shang\textsuperscript{*}}
\IEEEauthorblockA{\textit{Independent Researcher} \\
Frisco, USA \\
dingyanshang@gmail.com \\
\textsuperscript{*}Corresponding author}
\and
\IEEEauthorblockN{Zhenyu Xu}
\IEEEauthorblockA{\textit{Independent Researcher} \\
Fulshear, USA \\
zhenyuxu0918@gmail.com}
\and
\IEEEauthorblockN{Youting Wang}
\IEEEauthorblockA{\textit{Independent Researcher} \\
Mountain View, USA \\
wang.yout@northeastern.edu}
\linebreakand
\IEEEauthorblockN{Bonan Shen}
\IEEEauthorblockA{\textit{Independent Researcher} \\
Long Island City, USA \\
shenbonan2@gmail.com}
\and
\IEEEauthorblockN{Bowen Liu}
\IEEEauthorblockA{\textit{Independent Researcher} \\
South San Francisco, USA \\
bliu0962@usc.edu}
}

\maketitle

% =============================================================================
\begin{abstract}
Noise2Noise (N2N) trains denoisers on pairs of independently corrupted
observations, eliminating clean references. We stress-test two natural
conjectures about why the L1 loss outperforms L2 here. First, the
hypothesis that the L1 loss confers robustness via parameter
sparsity confuses the loss with Lasso regularization: an explicit
Lasso penalty produces the predicted sparsity yet fails to
reproduce L1's cross-noise behavior, while L1- and L2-trained
weight distributions are indistinguishable. Second, the population
optima of the two losses coincide exactly for symmetric signal
posteriors and nearly so for concentrated ones. Measured differences
are therefore dominated by optimization dynamics (bounded-influence
gradients), which we probe with gradient statistics and
contaminated-target training.
On Kodak24 with five synthetic noise families, the L1 loss holds a
statistically significant edge over L2, below 1 dB PSNR, holding
across three seeds on 13 of the 14 noise columns. On
real camera noise the loss is not the decisive variable in
distribution: on official
SIDD validation blocks, synthetic-Gaussian-trained N2N models gain only
0.8
to 3.7 dB over the noisy input regardless of loss, while retraining on
SIDD's own noisy pairs, never reading ground truth, gains 9.4 to
11.0 dB, far ahead of BM3D. All metrics are on raw network outputs,
and the study makes no leaderboard claim. The
training pair distribution, not the loss, carries the inductive
bias. That design rule applies wherever clean references are
unobtainable, from microscopy to industrial inspection sensors.
\end{abstract}

\begin{IEEEkeywords}
image denoising, self-supervised learning, Noise2Noise, robust statistics,
real-noise benchmarks, industrial inspection, nondestructive evaluation
\end{IEEEkeywords}

% =============================================================================
\section{Introduction}

Supervised deep
denoisers \cite{zhang2017dncnn,mao2016restoration} require registered clean/noisy pairs, costly or impossible in
microscopy, in low-light photography, and in industrial
nondestructive testing, where each unit under inspection is unique
and a noise-free reference of it is physically unobtainable
\cite{bagavathiappan2013thermography}. \NN{} (N2N) \cite{lehtinen2018noise2noise} showed that two
independent noisy observations of the same signal suffice: under zero-mean
noise, the network trained to map one observation to the other converges in
expectation to the clean-target minimizer.

Two practical questions remain. (i)~How much of N2N's observed
cross-noise robustness is attributable to the \emph{loss function}?
(ii)~The empirical edge of the L1 loss over L2 in restoration networks is
documented \cite{zhao2018loss} and attributed there to optimization
behavior; is that attribution right, or does the credit belong to
\emph{parameter sparsity}? We answer both with a controlled protocol
and make three contributions:

\begin{enumerate}
  \item \textbf{A corrected theoretical framing.} We distinguish the L1
  \emph{loss} (a residual penalty; maximum likelihood under a Laplace
  residual model) from L1 \emph{regularization} (a weight penalty; Lasso
  \cite{tibshirani1996lasso}), and we delimit what the loss can change:
  under zero-mean training noise the \MSE{} optimum is exactly the
  posterior mean, and the \MAE{} optimum coincides with it when the signal
  posterior is symmetric, and approximately when it is concentrated. The
  measurable loss-function effect reduces to finite-sample
  \emph{optimization dynamics} (\MAE{}'s bounded-influence gradients
  \cite{huber1981robust,zhao2018loss}), supported by gradient statistics
  and contaminated-target training.
  \item \textbf{A control experiment dissociating the L1 loss from L1
  regularization.} We train
  the same U-Net with \MSE{}, \MAE{}, and \MSE{}+Lasso under identical
  schedules. Only the explicit Lasso penalty produces weight sparsity, and
  it does \emph{not} inherit the \MAE{} cross-noise profile, ruling
  out sparsity as the carrier of L1's edge.
  \item \textbf{A loss-controlled quantification and closure of the
  real-noise domain gap.}
  On the official SIDD validation blocks \cite{abdelhamed2018sidd},
  synthetic-Gaussian-trained N2N recovers only $+0.8$ to $+3.7$~dB over
  the noisy input regardless of loss. Retraining the same five losses on SIDD-Medium
  noisy/noisy pairs (scene-disjoint 128/32 split, ground truth never read
  during training) lifts every variant to $+9.4$ to $+11.0$~dB, well
  ahead of BM3D \cite{dabov2007bm3d}. The gap itself is established
  knowledge \cite{guo2019cbdnet,brooks2019unprocessing}; new here is its
  same-architecture, same-budget measurement across five losses
  \emph{within} the N2N paradigm, its closure with noisy pairs only, and
  a content-matched control that splits the gap into ${\approx}5$~dB of
  scene statistics and ${\approx}3$~dB of noise distribution.
  This $\approx$8~dB regime gap dwarfs every in-distribution loss
  effect we measure ($\le$0.8~dB between \MAE{} and \MSE{} on synthetic
  noise, $\le$2.3~dB across the five real-pair losses); only out of
  distribution do loss effects grow large, governing robustness rather
  than accuracy (Section~\ref{sec:sidd}).
\end{enumerate}

All evaluations report PSNR, SSIM, and LPIPS \cite{zhang2018lpips} on the
\emph{raw} network output; a post-hoc image-enhancement stage present in
an earlier draft was removed as an evaluation confound
(Section~\ref{sec:enhancer}).

% =============================================================================
\section{Related Work}
\label{sec:related}

\textbf{Denoising without clean targets.}
N2N \cite{lehtinen2018noise2noise} trains on independently corrupted
pairs. Blind-spot methods remove even the second observation: \NV{}
\cite{krull2019noise2void} and Noise2Self \cite{batson2019noise2self}
predict masked pixels under a pixel-independence assumption;
Laine \emph{et al.} \cite{laine2019highquality} add noise-model-aware
posterior inference to the blind-spot architecture;
Noise2Same \cite{xie2020noise2same} relaxes strict $J$-invariance;
Noise2Sim \cite{niu2021noise2sim} mines self-similar patches;
Neighbor2Neighbor \cite{huang2021neighbor2neighbor},
Recorrupted-to-Recorrupted \cite{pang2021r2r}, and Noisier2Noise
\cite{moran2020noisier2noise} synthesize N2N-style pairs
from single noisy images; Self2Self \cite{quan2020self2self} trains on
one image alone with dropout ensembling; Blind2Unblind
\cite{wang2022blind2unblind}
revisits blind-spot information loss.

\textbf{Real-noise self-supervision.}
On real sRGB camera noise, spatial correlation breaks the pixel-independence
assumption of blind-spot training. That Gaussian-trained denoisers
transfer poorly to real photographs is itself established: CBDNet
\cite{guo2019cbdnet} motivates realistic noise modeling with this
failure, and Brooks \emph{et al.} \cite{brooks2019unprocessing} trace it
to the sRGB processing pipeline. AP-BSN \cite{lee2022apbsn} and CVF-SID
\cite{neshatavar2022cvfsid} are current self-supervised references on SIDD
(35.97 and 34.81~dB on the validation split; supervised state of the art
is 39--40~dB). More recent systems attack
precisely the mismatch we quantify, aligning the \emph{effective
training distribution} with the test noise: Positive2Negative
\cite{li2025p2n} replaces information-lossy masking and downsampling
with renoising-consistency training, next-scale prediction
\cite{shan2025nsp} builds cross-scale targets that decorrelate real
noise without destroying detail (37.1~dB), and Learning-to-Recorrupt
\cite{monroy2026l2r} learns the recorruption map when the noise
distribution is unknown. Our real-pair N2N results
(33--35~dB, Section~\ref{sec:sidd}) sit just below this self-supervised
band, which is consistent with our deliberately compact backbone and budget.
We do \emph{not} compete with these systems; the real-noise setting is a
controlled probe of whether the loss function or the training pair
distribution carries the inductive bias. Our N2N-on-real-pairs
configuration is simply \cite{lehtinen2018noise2noise} applied to
SIDD's two noisy shots per scene; its value is diagnostic. N2N
trained on SIDD-Medium pairs already appears as a baseline row in
the raw-domain tables of \cite{huang2021neighbor2neighbor}; we run the
configuration in sRGB, loss-controlled.

\textbf{Classical baselines.} BM3D \cite{dabov2007bm3d} remains the
reference non-learning denoiser and is still being refined, e.g.\ by
varying the transform in its collaborative-filtering stage
\cite{nurmana2024bm3d}; we therefore treat its configuration as an
experimental variable rather than a constant
(Section~\ref{sec:enhancer}).

\textbf{Loss functions for restoration.}
Zhao \emph{et al.} \cite{zhao2018loss} compare L2, L1, and SSIM-based
losses for restoration networks and attribute L1's edge over L2 to
optimization behavior rather than statistical modeling, consistent with
the framing we derive in Section~\ref{sec:theory}. Charbonnier
\cite{charbonnier1994} and Huber losses interpolate between the two
regimes.

% =============================================================================
\section{Theory: What the Loss Can and Cannot Do}
\label{sec:theory}

\subsection{Likelihood View and Its Limit}
\label{sec:likelihood}

For a residual model $r = x - f_\theta(x')$, minimizing
$\sum_i \rho(r_i)$ is maximum likelihood under $p(r) \propto e^{-\rho(r)}$:
\MSE{} corresponds to a Gaussian residual (optimal predictor: conditional
mean), \MAE{} to a Laplace one (conditional median).

In the N2N setting the training target is $x_2 = s + n_2$ with $n_2$
\emph{symmetric} zero-mean noise independent of $x_1$ \emph{and of the
signal} $s$: true by construction in our synthetic arm, violated by
signal-dependent real noise (Section~\ref{sec:sidd}). The \MSE{} optimum
is then exactly the posterior mean,
$\mathbb{E}[x_2 \mid x_1] = \mathbb{E}[s \mid x_1]$, with no further
assumptions. The \MAE{} optimum is the \emph{median} of
$p(x_2 \mid x_1)$, i.e., the signal posterior $p(s \mid x_1)$ convolved with
the noise density, and convolving a \emph{skewed} posterior with a
symmetric density does not symmetrize it. Hence, pointwise in the
predicted value $z$ given $x_1$,
\begin{equation}
\arg\min_{z}\, \mathbb{E}\!\left[(z - x_2)^2 \,\middle|\, x_1\right]
\;=\;
\arg\min_{z}\, \mathbb{E}\!\left[\,|z - x_2| \,\middle|\, x_1\right]
\label{eq:coincide}
\end{equation}
holds exactly when $p(s \mid x_1)$ is symmetric about its mean, and
approximately whenever the posterior is concentrated relative to the
noise, whose symmetric density then dominates the
convolution.\footnote{Clamping training pairs to $[0,1]$ truncates the
noise asymmetrically at saturated pixels, so the zero-mean-symmetry
premise is itself only approximate at the range bounds.}
Neither condition is guaranteed for natural images, so the
population-level gap between the two losses is the mean--median distance
of $p(x_2 \mid x_1)$: bounded above by one standard deviation of that
distribution, typically far smaller, but not zero.
Systematic differences between trained networks therefore combine
this small statistical component with the \emph{finite-sample
optimization path}: the \MAE{} gradient is $\mathrm{sign}(r)$,
so each residual's influence on the update is bounded
\cite{huber1981robust}, while
the \MSE{} gradient grows linearly in $r$, so rare large residuals
dominate updates. The sub-dB gaps we measure
(Section~\ref{sec:loss-effect}) are consistent with near-coincident
optima. We treat cross-noise evaluation as an empirical question.

\subsection{The L1-Loss-versus-L1-Regularization Confusion}
\label{sec:l1confusion}

Lasso regularization augments a base loss $J_0$ with a \emph{weight}
penalty, $J = J_0 + \alpha \sum_i |w_i|$,
whose fixed points exhibit sparse $w$ \cite{tibshirani1996lasso}. The
\emph{loss} $J_0 = \|x_2 - f_\theta(x_1)\|_1$ penalizes pixel-domain
residuals; its gradient flows to activations, not weights, and no
mechanism connects it to weight sparsity.
Section~\ref{sec:weighthist} tests the dissociation directly.

\subsection{Hypotheses}
\label{sec:hypothesis}

(H1)~\MAE{}-class losses (\MAE{}, Charbonnier, Huber) generalize no worse
than \MSE{} to additive heavy-tailed test noise, with any advantage
concentrated in perceptual metrics.
(H2)~Weight sparsity is neither necessary nor sufficient for the
\MAE{}-class cross-noise profile.
(H3)~When the test noise violates the additive zero-mean training
assumption structurally (multiplicative speckle at high variance; real
signal-dependent sRGB camera noise), N2N fails \emph{irrespective of loss},
and matching the training pair distribution to the test noise
restores the guarantee.

% =============================================================================
\section{Method}

\subsection{Architecture}
A U-Net \cite{ronneberger2015unet} (depth~4, base width~48, BN, ReLU) with
a \emph{global residual connection}: the network predicts a correction
added to its input, DnCNN-style \cite{zhang2017dncnn}; held fixed across
all loss variants.

\subsection{Training Distributions}
\label{sec:traindist}
\textbf{Synthetic arm.} 400 color images from the BSDS500 train+test
splits (the standard CBSD400 set) \cite{martin2001bsds} (disjoint from
all evaluation sets). Each
sample is a random $128{\times}128$ crop; $x_1, x_2$ receive independent
additive Gaussian noise with $\sigma_1, \sigma_2 \sim \mathcal{U}(5, 50)$
(8-bit scale). The training family is deliberately narrow so that
cross-noise behavior is attributable to the loss.
\textbf{Real arm.} SIDD-Medium sRGB: 160 scenes, two camera shots each.
Scenes are split 128/32 by sorted scene-instance name (deterministic,
seed-free). Training draws a common random crop from the two
\emph{full-resolution} noisy shots of a training scene; ground truth is
never read. We verified N2N's cross-shot independence premise:
over all 160 scenes the Pearson correlation between the two shots'
residuals (noisy $-$ GT, analysis only) has median 0.026 (mean 0.058;
11 scenes above 0.2); any shared component is \emph{retained} by N2N.
In contrast, the \emph{within}-shot residual is strongly spatially
correlated (lag-1 autocorrelation ${\approx}0.46$ in both directions):
the pixel-independence violation that handicaps blind-spot
methods on real sRGB noise. All training and evaluation is at native
resolution: bicubic downsampling (used in an earlier draft) partially
averages out the very noise the network must learn
(Section~\ref{sec:enhancer}).

\subsection{Losses}
\MSE{}; \MAE{}; Charbonnier ($\varepsilon{=}10^{-3}$)
\cite{charbonnier1994}; Huber (Smooth-L1, $\delta{=}0.05$ at the
residual scale; the conventional $\delta{=}1$ exceeds every residual
on $[0,1]$ data); and
\MSE{}+Lasso ($\alpha{=}10^{-5}$ on conv weights) as the sparsity control.
Budgets are matched everywhere: 50 epochs $\times$ 5000 crops, batch~16,
Adam $10^{-4}$, cosine decay, identical for synthetic and real arms
($\approx$15.6k steps), removing training-budget confounds.

\subsection{Evaluation Protocol}
\textbf{Synthetic:} Kodak24 at full resolution; noise grid: Gaussian
$\sigma \in \{15,25,50\}$; random-impulse (each corrupted pixel--channel
replaced by a $\mathcal{U}(0,1)$ draw) $p \in \{0.1,0.3,0.6\}$;
salt-and-pepper ($0$/$1$ extremes, equiprobable) $p \in \{0.05,0.1\}$;
Poisson ($y = \mathrm{Pois}(\lambda x)/\lambda$, $\lambda$ the expected
count at unit intensity) $\lambda \in \{30,60\}$;
speckle $v \in \{0.05,0.1,0.2,0.5\}$ (the two upper levels probe the H3
collapse regime).
The five families span the noise regimes of
deployed imaging sensors: Gaussian read noise; impulse and
salt-and-pepper defective-pixel artifacts characteristic of
microbolometer (thermal) arrays; Poisson counting noise of
photon-limited radiographic and short-wave-infrared capture; and
multiplicative speckle of coherent and ultrasonic imaging, a target of
dedicated despeckling models \cite{patil2025speckle}.
\textbf{Real:} (i)~the official SIDD validation blocks (40 images
$\times$ 32 blocks of $256^2$), directly comparable to the literature;
(ii)~the held-out 32 scenes of our split at native resolution (tiled
inference), scene-disjoint from training by construction.
\textbf{Metrics:} PSNR, SSIM, LPIPS (AlexNet, \texttt{lpips}~0.1.4) on
raw outputs (LPIPS is
unbounded above; heavily corrupted inputs can exceed 1).
Paired Wilcoxon signed-rank tests over per-image scores, Holm-corrected
across all cells, accompany the key \MSE{}-vs-\MAE{} comparisons. The
unit of analysis is one denoised image: the 32 blocks of each official
validation image are averaged into a single per-image score before
testing ($n{=}40$), and on the held-out scenes each of the two shots
per scene is denoised and scored separately ($n{=}64$);
significance statements are conditional on the single trained model per
cell, with seed replication (Section~\ref{sec:loss-effect}) as the
partial guard.
\textbf{Baselines:} color BM3D (CBM3D, \texttt{bm3d}~4.0.3 with
\texttt{bm3d\_rgb}; given the true $\sigma$ on Gaussian columns, a
wavelet-domain MAD estimate otherwise) and \NV{}, a compact
reimplementation (stratified masking, 1.5\% of pixels, $11{\times}11$
replacement window) meant as a fair external reference rather than a tuned
system, trained twice under the same budget:
on the synthetic images, and directly on the real SIDD noisy shots
(single-image blind-spot, ground truth never read), so each training
regime has a blind-spot reference.

% =============================================================================
\section{Experiments}
\label{sec:experiments}

\subsection{Cross-Noise Generalization}
\label{sec:crosstable}

% Auto-generated from results/cross_noise_kodak_v4.json by render_tables.py (layout=compact).
% Re-run to refresh after a new evaluation pass.

\begin{table*}[t]
\centering\scriptsize\setlength{\tabcolsep}{3pt}
\caption{Cross-noise generalization, PSNR (dB, higher is better). Bold marks the best method per column (noisy input excluded). Top block: non-learning (BM3D) and self-supervised (Noise2Void) baselines. Middle block: Noise2Noise trained on synthetic Gaussian noise with five losses. Bottom block: models trained on SIDD-Medium real noise without ground truth---Noise2Void on single noisy shots, the five Noise2Noise losses on noisy/noisy pairs. Column parameters differ by family and are not a common percentage. $\sigma$: Gaussian s.d.\ in 8-bit levels ($25$ is $25/255$). imp, sp: percent of pixel values corrupted, drawn per channel (uniform draws; 0 or 1 at 50/50). poi: Poisson rate $\lambda$, $y{=}\mathrm{Pois}(\lambda x)/\lambda$, so larger is \emph{less} noise. spk: speckle variance $\times$100 in $y{=}x{+}xn$, $n\sim\mathcal{N}(0,v)$ ($50$ is $v{=}0.5$). SIDD: 32 held-out scenes at native resolution; SIDD val: official validation blocks.}
\label{tab:cross-psnr}
\begin{tabular}{lrrrrrrrrrrrrrrrr}
\toprule
Method & \multicolumn{1}{c}{$\sigma{=}15$} & \multicolumn{1}{c}{$\sigma{=}25$} & \multicolumn{1}{c}{$\sigma{=}50$} & \multicolumn{1}{c}{imp\,10} & \multicolumn{1}{c}{imp\,30} & \multicolumn{1}{c}{imp\,60} & \multicolumn{1}{c}{sp\,5} & \multicolumn{1}{c}{sp\,10} & \multicolumn{1}{c}{poi\,30} & \multicolumn{1}{c}{poi\,60} & \multicolumn{1}{c}{spk\,5} & \multicolumn{1}{c}{spk\,10} & \multicolumn{1}{c}{spk\,20} & \multicolumn{1}{c}{spk\,50} & \multicolumn{1}{c}{SIDD} & \multicolumn{1}{c}{SIDD\,val} \\
\cmidrule(lr){2-15} \cmidrule(lr){16-17}
\emph{noisy input} & 24.76 & 20.44 & 14.87 & 18.57 & 13.81 & 10.80 & 18.14 & 15.13 & 18.99 & 21.92 & 20.27 & 17.40 & 14.71 & 11.81 & 27.37 & 23.66 \\
\midrule
BM3D & \textbf{34.25} & \textbf{31.47} & \textbf{27.34} & 20.23 & 20.97 & 16.80 & 18.56 & 17.61 & 29.43 & 31.36 & 26.60 & 24.30 & 22.32 & 20.73 & 28.48 & 24.89 \\
Noise2Void & 29.15 & 27.94 & 25.23 & 26.19 & 20.98 & 16.36 & 26.96 & 24.87 & 27.26 & 28.33 & 27.48 & 26.06 & 24.29 & 21.59 & 32.64 & 29.55 \\
\midrule
N2N-MSE & 32.44 & 30.27 & 26.64 & 27.59 & 21.89 & 16.71 & 27.73 & 26.05 & 29.36 & 31.01 & 29.81 & 27.70 & 25.22 & 21.79 & 30.23 & 26.48 \\
\;\;+ Lasso & 32.48 & 30.10 & 25.99 & 27.60 & 21.51 & 16.51 & \textbf{28.09} & 25.62 & 29.02 & 30.89 & 29.55 & 27.39 & 25.02 & 22.02 & 30.98 & 27.38 \\
N2N-MAE & 33.22 & 30.68 & 26.97 & 28.02 & 22.30 & 16.85 & 27.82 & 26.31 & \textbf{29.71} & \textbf{31.52} & \textbf{30.36} & 28.08 & 25.53 & 22.05 & 29.65 & 25.86 \\
N2N-Charb. & 32.39 & 30.65 & 26.95 & 28.01 & 22.22 & 16.78 & 27.79 & 26.24 & 29.69 & 31.24 & 29.66 & \textbf{28.10} & 25.54 & 22.19 & 28.20 & 24.45 \\
N2N-Huber & 32.86 & 30.60 & 26.90 & \textbf{28.03} & \textbf{22.30} & \textbf{16.87} & 27.96 & \textbf{26.34} & 29.62 & 31.43 & 30.24 & 28.08 & \textbf{25.64} & \textbf{22.33} & 27.81 & 25.29 \\
\midrule
Noise2Void (real) & 23.52 & 20.52 & 15.46 & 18.74 & 12.44 & 7.49 & 18.33 & 14.31 & 19.07 & 21.55 & 20.28 & 18.13 & 14.64 & 9.03 & 28.56 & 25.44 \\
N2N-MSE (real) & 26.65 & 24.77 & 21.03 & 22.66 & 18.18 & 14.43 & 22.67 & 20.45 & 23.80 & 25.33 & 24.34 & 22.52 & 20.39 & 17.56 & 34.72 & 34.46 \\
\;\;+ Lasso (real) & 25.90 & 24.36 & 20.62 & 22.49 & 17.96 & 14.33 & 22.74 & 20.45 & 23.58 & 24.92 & 24.04 & 22.53 & 20.72 & 18.08 & 34.27 & 33.09 \\
N2N-MAE (real) & 26.92 & 23.85 & 11.22 & 19.36 & 11.70 & 8.01 & 18.92 & 12.33 & 23.24 & 25.21 & 24.08 & 21.96 & 16.81 & 7.15 & \textbf{35.81} & \textbf{34.61} \\
N2N-Charb. (real) & 26.87 & 24.39 & 20.14 & 22.01 & 17.46 & 13.85 & 21.93 & 19.59 & 23.28 & 25.22 & 24.10 & 22.02 & 19.85 & 17.24 & 33.56 & 33.79 \\
N2N-Huber (real) & 26.77 & 24.47 & 20.35 & 22.17 & 17.59 & 13.94 & 22.12 & 19.79 & 23.50 & 25.29 & 24.31 & 22.32 & 20.09 & 17.28 & 34.12 & 34.00 \\
\bottomrule
\end{tabular}
\end{table*}

\begin{table*}[t]
\centering\scriptsize\setlength{\tabcolsep}{2.5pt}
\caption{SSIM and LPIPS (AlexNet) on representative columns. Conventions as in Table~\ref{tab:cross-psnr}.}
\label{tab:cross-perceptual}
\begin{tabular}{lrrrrrrrr@{\hspace{6pt}}rrrrrrrr}
\toprule
 & \multicolumn{8}{c}{SSIM $\uparrow$} & \multicolumn{8}{c}{LPIPS $\downarrow$} \\
\cmidrule(lr){2-9} \cmidrule(lr){10-17}
Method & \multicolumn{1}{c}{$\sigma{=}25$} & \multicolumn{1}{c}{imp\,30} & \multicolumn{1}{c}{sp\,10} & \multicolumn{1}{c}{poi\,60} & \multicolumn{1}{c}{spk\,10} & \multicolumn{1}{c}{spk\,50} & \multicolumn{1}{c}{SIDD} & \multicolumn{1}{c}{SIDD\,val} & \multicolumn{1}{c}{$\sigma{=}25$} & \multicolumn{1}{c}{imp\,30} & \multicolumn{1}{c}{sp\,10} & \multicolumn{1}{c}{poi\,60} & \multicolumn{1}{c}{spk\,10} & \multicolumn{1}{c}{spk\,50} & \multicolumn{1}{c}{SIDD} & \multicolumn{1}{c}{SIDD\,val} \\
\midrule
\emph{noisy input} & 0.350 & 0.146 & 0.215 & 0.438 & 0.329 & 0.147 & 0.551 & 0.328 & 0.563 & 1.100 & 0.944 & 0.453 & 0.704 & 1.110 & 0.662 & 0.792 \\
\midrule
BM3D & \textbf{0.867} & \textbf{0.573} & 0.287 & 0.867 & 0.679 & \textbf{0.554} & 0.610 & 0.384 & 0.178 & \textbf{0.403} & 0.728 & 0.143 & 0.317 & 0.483 & 0.612 & 0.744 \\
Noise2Void & 0.759 & 0.543 & 0.624 & 0.783 & 0.706 & 0.497 & 0.824 & 0.628 & 0.266 & 0.520 & 0.405 & 0.236 & 0.328 & 0.532 & 0.284 & 0.490 \\
\midrule
N2N-MSE & 0.834 & 0.541 & 0.630 & 0.864 & 0.769 & 0.499 & 0.685 & 0.444 & 0.145 & 0.424 & 0.330 & 0.119 & 0.210 & 0.488 & 0.516 & 0.681 \\
\;\;+ Lasso & 0.814 & 0.527 & 0.608 & 0.849 & 0.743 & 0.500 & 0.741 & 0.496 & 0.187 & 0.491 & 0.385 & 0.145 & 0.262 & 0.513 & 0.424 & 0.616 \\
N2N-MAE & 0.839 & 0.552 & 0.643 & \textbf{0.868} & 0.772 & 0.510 & 0.658 & 0.422 & \textbf{0.144} & 0.417 & \textbf{0.323} & \textbf{0.117} & \textbf{0.208} & 0.467 & 0.553 & 0.710 \\
N2N-Charb. & 0.838 & 0.551 & 0.637 & 0.863 & 0.773 & 0.511 & 0.659 & 0.417 & 0.148 & 0.417 & 0.329 & 0.121 & 0.209 & \textbf{0.466} & 0.534 & 0.704 \\
N2N-Huber & 0.839 & 0.554 & \textbf{0.645} & 0.868 & \textbf{0.774} & 0.511 & 0.639 & 0.415 & 0.147 & 0.420 & 0.325 & 0.120 & 0.213 & 0.477 & 0.555 & 0.709 \\
\midrule
Noise2Void (real) & 0.340 & 0.137 & 0.177 & 0.405 & 0.286 & 0.174 & 0.625 & 0.418 & 0.473 & 0.878 & 0.776 & 0.392 & 0.592 & 0.777 & 0.552 & 0.637 \\
N2N-MSE (real) & 0.511 & 0.205 & 0.266 & 0.565 & 0.424 & 0.246 & 0.878 & \textbf{0.831} & 0.430 & 0.777 & 0.666 & 0.375 & 0.499 & 0.757 & 0.230 & \textbf{0.163} \\
\;\;+ Lasso (real) & 0.519 & 0.233 & 0.291 & 0.563 & 0.439 & 0.272 & 0.852 & 0.797 & 0.443 & 0.761 & 0.663 & 0.404 & 0.542 & 0.750 & 0.285 & 0.269 \\
N2N-MAE (real) & 0.474 & 0.131 & 0.164 & 0.558 & 0.401 & 0.039 & \textbf{0.879} & 0.798 & 0.448 & 0.866 & 0.798 & 0.363 & 0.522 & 1.025 & \textbf{0.225} & 0.192 \\
N2N-Charb. (real) & 0.492 & 0.178 & 0.234 & 0.560 & 0.404 & 0.214 & 0.832 & 0.780 & 0.436 & 0.842 & 0.731 & 0.366 & 0.524 & 0.797 & 0.249 & 0.200 \\
N2N-Huber (real) & 0.500 & 0.187 & 0.244 & 0.567 & 0.412 & 0.220 & 0.845 & 0.794 & 0.444 & 0.834 & 0.719 & 0.377 & 0.522 & 0.788 & 0.247 & 0.198 \\
\bottomrule
\end{tabular}
\end{table*}

Table~\ref{tab:cross-psnr} reports the full grid; Table~\ref{tab:cross-perceptual}
gives SSIM/LPIPS on representative columns. Four observations.
(i)~\emph{In-distribution Gaussian}: CBM3D, given the true $\sigma$,
leads every compact-budget N2N variant: 31.47 vs.\ 30.1--30.7~dB at
$\sigma{=}25$, by 1.0--1.9~dB at $\sigma{=}15$, and by 0.4--1.4~dB at
$\sigma{=}50$.
Beating a tuned classical baseline in distribution was never the aim of
this budget; the interest is off-distribution.
(ii)~\emph{Heavy-tailed noise}: the N2N variants lead BM3D by 7--9.5~dB
on 10\% impulse and both salt-and-pepper columns, where the Gaussian
noise model and the white-noise $\sigma$-estimate break down; at 30\%
impulse the margin narrows to 0.5--1.3~dB, and at 60\% every method
collapses below 17~dB (best learned 16.87 vs.\ BM3D 16.80).
(iii)~\emph{Poisson}: the five losses agree to within 0.7~dB, as the
Anscombe view predicts (mild Poisson is a rescaled Gaussian); the
residual tails carry no signal to exploit, and CBM3D sits inside the
same band.
(iv)~\emph{Speckle}: contrary to our initial H3 expectation of a collapse,
additive-Gaussian-trained N2N still gains $+10.0$ to $+10.5$~dB over the
noisy input at $v{=}0.5$. The parameterization explains why. Our speckle
$y = x + x\cdot n$, $n \sim \mathcal{N}(0, v)$, is conditionally additive
zero-mean Gaussian with signal-proportional standard deviation
$x\sqrt{v}$, which lies inside the trained family with spatially modulated
variance, so the assumption degrades gracefully rather than
failing structurally.
The graceful mode is practically relevant, as multiplicative speckle is
the dominant noise family of coherent modalities, ultrasonic and
acoustic imaging among them.

\subsection{The Loss-Function Effect}
\label{sec:loss-effect}

Among the four pure losses, \MAE{} leads \MSE{} on
\emph{every} synthetic column of Table~\ref{tab:cross-psnr} (seed~0):
$+0.09$ to $+0.78$~dB PSNR and $+0.002$--$0.013$ SSIM. Paired Wilcoxon tests over the 24 Kodak images,
Holm-corrected across all 45 (noise, metric) cells (the 14 synthetic
columns plus the native-resolution SIDD set, each under PSNR, SSIM, and
LPIPS), confirm the PSNR
advantage on 13 of 14 columns (salt-and-pepper~5\% is marginal: raw
$p=0.004$, corrected $0.06$) and the SSIM advantage on 11 of 14; the
LPIPS advantage survives correction only under the heaviest corruption
(speckle~0.5).

% --- BEGIN seed-gap table. To drop it for length: delete lines to END,
% delete " (Table~\ref{tab:seed-gap})" below, and in Sec. Reproducibility
% change "so Table~\ref{tab:seed-gap} recomputes" -> "so that comparison
% recomputes". The prose then still answers the seed question on its own. ---
% Auto-generated from results/kodak_seeds_v4.json by render_tables.py.
% Re-run to refresh after a new seed pass.

\begin{table*}[t]
\centering\scriptsize\setlength{\tabcolsep}{3pt}
\caption{\MAE{}$-$\MSE{} PSNR gap (dB) per training seed over the full synthetic grid; positive favors \MAE{}. The gap holds under all three seeds on 13 of 14 columns and all five noise families; the exception is the heaviest speckle level, where one seed reverses the sign. Parameters as in Table~\ref{tab:cross-psnr}.}
\label{tab:seed-gap}
\begin{tabular}{lrrrrrrrrrrrrrr}
\toprule
Seed & \multicolumn{1}{c}{$\sigma{=}15$} & \multicolumn{1}{c}{$\sigma{=}25$} & \multicolumn{1}{c}{$\sigma{=}50$} & \multicolumn{1}{c}{imp\,10} & \multicolumn{1}{c}{imp\,30} & \multicolumn{1}{c}{imp\,60} & \multicolumn{1}{c}{sp\,5} & \multicolumn{1}{c}{sp\,10} & \multicolumn{1}{c}{poi\,30} & \multicolumn{1}{c}{poi\,60} & \multicolumn{1}{c}{spk\,5} & \multicolumn{1}{c}{spk\,10} & \multicolumn{1}{c}{spk\,20} & \multicolumn{1}{c}{spk\,50} \\
\cmidrule(lr){2-15}
0 & +0.78 & +0.41 & +0.33 & +0.42 & +0.41 & +0.14 & +0.09 & +0.26 & +0.34 & +0.51 & +0.54 & +0.38 & +0.31 & +0.26 \\
1 & +0.88 & +0.48 & +0.32 & +0.46 & +0.33 & +0.14 & +0.17 & +0.22 & +0.36 & +0.57 & +0.53 & +0.38 & +0.20 & -0.01 \\
2 & +0.92 & +0.59 & +0.46 & +0.46 & +0.24 & +0.08 & +0.18 & +0.20 & +0.45 & +0.62 & +0.56 & +0.32 & +0.07 & +0.42 \\
\midrule
mean & +0.86 & +0.49 & +0.37 & +0.45 & +0.33 & +0.12 & +0.15 & +0.23 & +0.39 & +0.57 & +0.54 & +0.36 & +0.19 & +0.22 \\
s.d. & 0.07 & 0.09 & 0.07 & 0.02 & 0.09 & 0.03 & 0.05 & 0.03 & 0.06 & 0.06 & 0.02 & 0.04 & 0.12 & 0.22 \\
\bottomrule
\end{tabular}
\end{table*}

% --- END seed-gap table ---
\textbf{Seed replication beyond the Gaussian columns.} We retrained
\MAE{} and \MSE{} under two further seeds and re-ran the \emph{entire}
14-column grid for each (Table~\ref{tab:seed-gap}). The gap survives
all three seeds on 13 of the 14 columns and all five noise families;
the exception is the heaviest speckle level ($v{=}0.5$), where one seed
reverses the sign and the deviation runs some four times the typical
column's, so there the loss effect
is no longer separable from seed noise. It is the endpoint of a trend,
not an isolated outlier: across the speckle family the seed spread
rises monotonically with the variance ($0.02\rightarrow0.22$~dB) as the
mean gap more than halves, so heavier corruption erodes the loss effect
faster than the seed effect. The largest single-seed gap
anywhere is $+0.92$~dB. The remaining three losses were seed-replicated
in the real-pair cell instead (Section~\ref{sec:sidd}), so the seed-0
ordering at $\sigma{=}25$ (\MAE{} $>$ Charbonnier $>$ Huber $>$ \MSE{}
$>$ Lasso) is a
single-seed observation we do not claim to be seed-stable.
Charbonnier tracks \MAE{} closely (except at
$\sigma{=}15$, where \MAE{}'s margin is largest), and Huber at
$\delta{=}0.05$ interpolates as
intended: 30.60~dB at $\sigma{=}25$, between \MSE{} (30.27) and \MAE{}
(30.68), siding with \MAE{} on the heavy-tailed columns, whereas at
the conventional $\delta{=}1$ it never leaves the quadratic regime and
replicates \MSE{} (30.34~dB; not tabulated). H1's core survives, but its rider is
refuted: the advantage concentrates in PSNR and SSIM and largely
vanishes in LPIPS (the opposite of the predicted perceptual
concentration), and it remains an order of magnitude smaller than
the training-distribution effects of Section~\ref{sec:sidd}.

\textbf{Probing the mechanism.} Two diagnostics separate the
optimization-dynamics account from the (near-empty) statistical one of
\eqref{eq:coincide}.
Per-step gradient norms differ only modestly at the batch level
(max/median 3.1 under \MAE{} vs.\ 4.0 under \MSE{}; CV 0.35 vs.\ 0.39):
averaging over ${\sim}10^6$ pixels per step conceals per-pixel
influence. The second diagnostic is behavioral: replacing 5\% of
\emph{training-target} pixels with salt-and-pepper extremes costs the
\MSE{} model 0.5--1.7~dB on the Gaussian columns ($-1.04$ at
$\sigma{=}25$) but the \MAE{} model at most $-0.6$ ($-0.07$ at
$\sigma{=}25$): contamination that drags the conditional mean leaves the
conditional median nearly unchanged, as the bounded-influence account
predicts.

One asymmetry: on the real-noise columns the ordering
\emph{reverses} for synth-trained models (\MSE{} beats \MAE{} by
0.62~dB on the official validation blocks, $n{=}40$ images,
Holm-corrected over the three blocks-set metrics, $p<10^{-4}$; the held-out 32-scene set agrees,
$+0.58$~dB, $n{=}64$). SSIM (0.444 vs.\ 0.422) and LPIPS (0.681
vs.\ 0.710) side with \MSE{} as well, so the reversal is not an artifact of
PSNR's alignment with \MSE{}. Bounded-influence training sharpens
specialization to the training tails; it does not buy general
out-of-distribution robustness. Note also that on real noise the
near-coincidence argument of Section~\ref{sec:likelihood} no longer
applies: the conditional noise is signal-dependent and clipped, so the
mean--median gap of $p(x_2 \mid x_1)$ is genuinely non-zero and PSNR
structurally rewards the conditional mean. The optimization-dynamics
account is thus delimited to the symmetric synthetic setting; on the
real arm this statistical component is a live co-explanation.

\subsection{Weight Histograms: the Sparsity Hypothesis Fails}
\label{sec:weighthist}

\begin{figure}[t]
\centering
\includegraphics[width=0.95\linewidth]{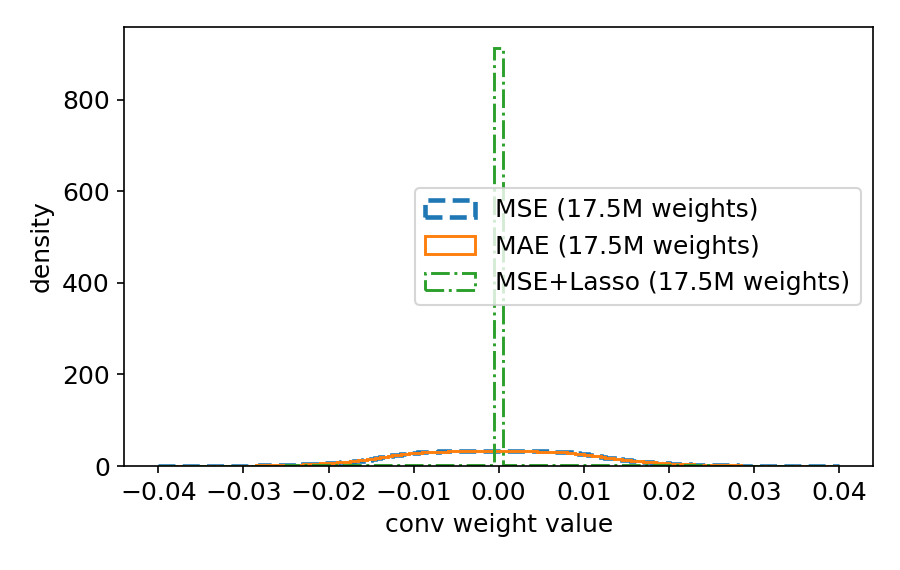}
\caption{Conv-weight histograms, identical schedules: \MSE{}, \MAE{},
\MSE{}+Lasso. Only the explicit weight penalty produces a zero-spike
(91.3\% of weights within $10^{-3}$ of zero vs.\ 6.6\%/6.5\%); the \MSE{}
(dashed) and \MAE{} (solid) curves coincide.}
\label{fig:weighthist}
\end{figure}

The \MSE{} and \MAE{} histograms coincide
(Fig.~\ref{fig:weighthist}): 6.6\% and 6.5\% of the 17.5M
conv weights lie within $10^{-3}$ of zero, with matching spread
($\mathrm{std}$ 0.0112 vs.\ 0.0115). The Lasso model concentrates 91.3\%
of its weights there ($\mathrm{std}$ 0.0053) yet tracks \MSE{} across
the noise grid (Table~\ref{tab:cross-psnr}), showing none of the \MAE{}
profile. The sparse
model is not capacity-crippled (it stays within 0.2~dB of \MSE{} at
$\sigma{=}25$, 30.10 vs.\ 30.27~dB), nor is the regime a knife-edge
artifact of $\alpha$: across two decades ($10^{-6}$--$10^{-4}$) the
near-zero fraction saturates at ${\approx}91\%$ and $\sigma{=}25$ PSNR
stays \MSE{}-like (30.38/30.10/30.29~dB). Sparsity behaves exactly as
Section~\ref{sec:l1confusion} predicts: easy to induce, and it never brings
the \MAE{} profile with it. H2 is confirmed.

\subsection{The Real-Noise Domain Gap, and Closing It}
\label{sec:sidd}

\begin{figure*}[t]
\centering
\includegraphics[width=0.96\textwidth]{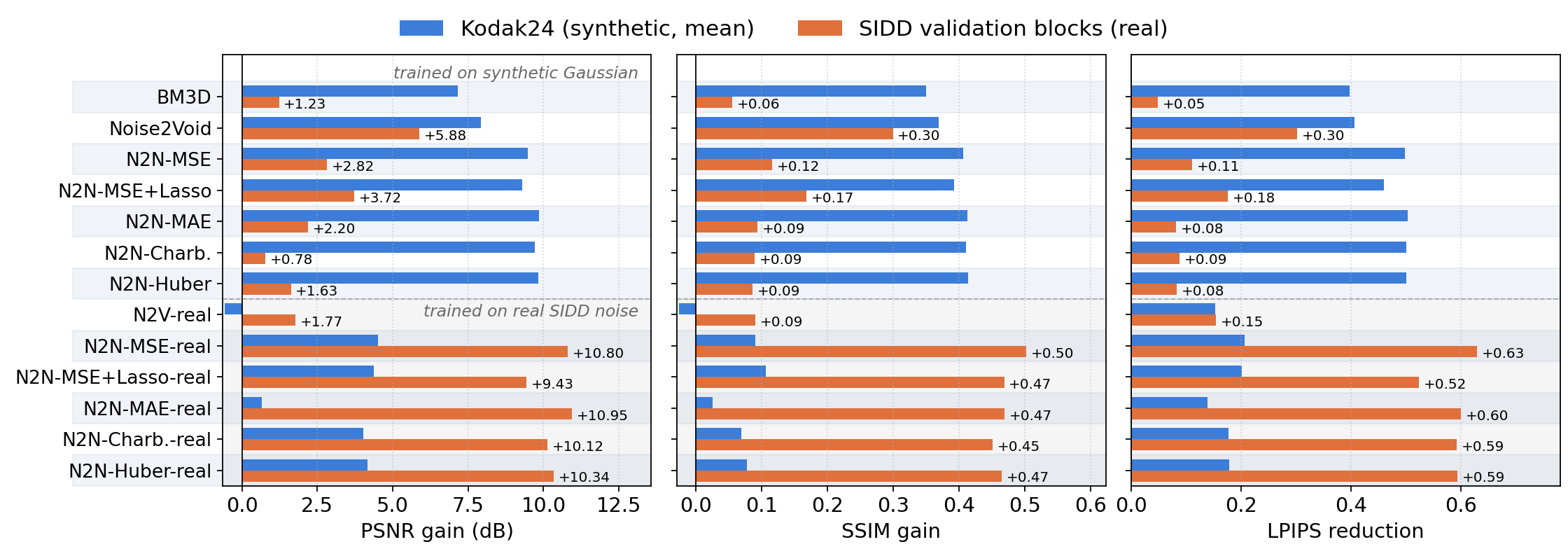}
\caption{Metric gain over the noisy input of
Table~\ref{tab:cross-psnr}: Kodak24 average (blue) vs.\
official SIDD blocks (orange). Above the separator: BM3D ($+1.2$~dB on
real noise), \NV{} ($+5.9$), and the five
synth-trained N2N variants ($+0.8$ to $+3.7$~dB). Below (shaded):
real-noise-trained: \NV{} gains only $+1.8$~dB; N2N on noisy pairs,
$+9.4$ to $+11.0$~dB, at the cost of synthetic performance.}
\label{fig:sidd}
\end{figure*}

\textbf{Synth-trained N2N recovers only a fraction of the available
real-noise gain.} On the official validation blocks (noisy input: 23.66~dB PSNR),
the five Gaussian-trained variants score 24.5--27.4~dB
(Fig.~\ref{fig:sidd}): a $+0.8$ to
$+3.7$~dB improvement, loss-independent in kind, well below \NV{}
(29.6~dB) and what the data supports; LPIPS agrees
(0.62--0.71 vs.\ noisy 0.79). Blind-spot \NV{} trained on the
same Gaussian images in fact transfers \emph{better} than N2N here, since masking
forces it to model context rather than the training residual.
In the converse experiment, \NV{} trained directly
on the real SIDD shots gains only $+1.8$~dB (25.44~dB on the blocks),
\emph{below} its synth-trained counterpart, and acts as a
near-identity on Kodak. Because the within-shot residual is spatially
correlated (Section~\ref{sec:traindist}, lag-1 ${\approx}0.46$), the
masked pixel's noise is predictable from its neighbors, and the
blind-spot objective reproduces it. Real-pair N2N, structurally immune,
gains $+10.3$~dB on average from the same data and budget.

\textbf{Real-pair training closes the gap.} Retrained on
native-resolution SIDD noisy pairs, the same five losses reach
33.1--34.6~dB ($+9.4$ to $+11.0$~dB over the input), 8.2--9.7~dB ahead
of BM3D (24.9~dB: its white-noise $\sigma$-estimate underestimates the
spatially correlated SIDD noise of Section~\ref{sec:traindist}, a
failure mode shared by any i.i.d.-noise assumption), just below the
published self-supervised band. SSIM rises from 0.33 (noisy) to 0.78--0.83, LPIPS falls from
0.79 to 0.16--0.27. The cross-loss spread (1.5~dB at seed~0) is small
against the
$\approx$8~dB separation between the two training regimes, and seed
replication shows much of it is noise: retraining all five real-pair
losses with two further seeds gives per-loss standard deviations of
0.16--0.80~dB on the blocks (vs.\ $\le$0.06~dB for \MAE{} and \MSE{}
in the synthetic cell),
a 1.4~dB spread of per-loss means, and a within-column ordering that is
not seed-stable (only Lasso is consistently last). In distribution,
loss choice among real-pair models is second-order; the fix is
carried by the pair distribution. The scene-disjoint
32-scene column at native resolution (Table~\ref{tab:cross-psnr}, SIDD)
agrees: real-pair variants 33.6--35.8~dB vs.\ synth-trained
27.8--31.0~dB.

\textbf{Content or noise?} The two regimes differ in scene content as
well as noise, so we ran a diagnostic control: N2N trained on SIDD
ground-truth scenes with synthetic Gaussian pairs (clean images used
only as the base for synthetic corruption, exactly as CBSD400 is in the
synthetic arm; this control is \emph{not} self-supervised). It reaches
31.50~dB on the blocks: of the 7.98~dB \MSE{} regime gap, scene
statistics account for ${\approx}5.0$~dB and matching the sensor's
actual noise for the remaining ${\approx}3.0$~dB. This is a one-path
decomposition, holding the synthetic noise family fixed while swapping
content; the content--noise interaction term is not identified, so the
split should not be read as a variance decomposition. Both components are
real; only the noisy-pair route captures both without any clean data.
A supervised anchor completes the comparison: the same U-Net at
the same budget trained noisy$\to$ground-truth (N2C; again not
self-supervised) reaches 36.18~dB on the blocks, 1.3~dB above the best
real-pair per-loss mean over three seeds. The noisy-target
concession is small next to the ${\approx}8$~dB pair-distribution
effect, and the anchor's cross-noise profile specializes exactly like
the real-pair family.

\begin{figure*}[t]
\centering
\includegraphics[width=0.96\textwidth]{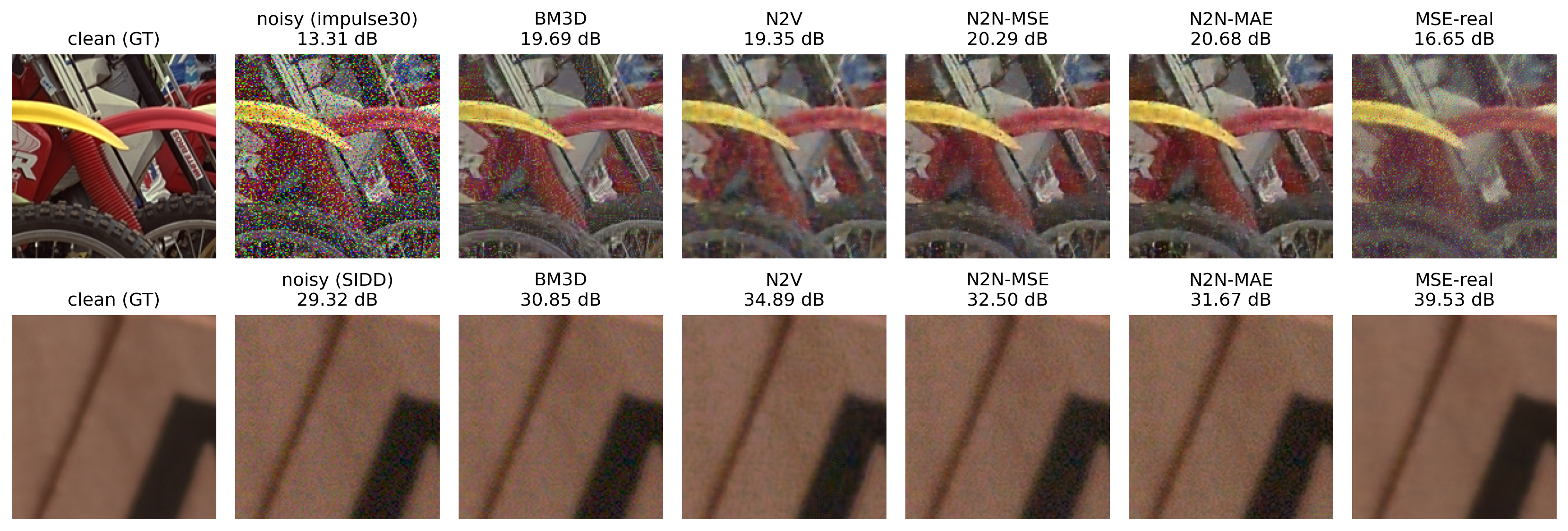}
\caption{Zoomed crops, raw outputs, per-crop PSNR inset; the \NV{}
column is the synth-trained variant in both rows. Top: Kodak24,
30\% random impulse noise: synth-trained variants suppress impulses
that BM3D's Gaussian model cannot; the real-pair model (right) fails on
unseen noise. Bottom: held-out SIDD scene, native resolution, where the
relation inverts; only the real-pair model fully cleans the camera
noise.}
\label{fig:qual}
\end{figure*}

\textbf{Failure modes of real-pair models.} Real-pair models transfer poorly \emph{back} to
heavy synthetic noise: at $\sigma{=}50$ they lose 5.4--6.8~dB to their
synth-trained counterparts, and \MAE{}-real degenerates outright on
far-out-of-distribution columns (11.2~dB at $\sigma{=}50$, 7.2~dB on
speckle~0.5, below the noisy input). The degenerate mode is
saturation rather than over-smoothing: at $\sigma{=}50$ \MAE{}-real rails
19\% of output pixels to the $[0,1]$ bounds (58\% under speckle~0.5),
while \MSE{}-real merely over-smooths and degrades gracefully;
median-trained corrections carry no penalty proportional to overshoot. The
brittleness is specific to the pure median objective: Charbonnier-real
and Huber-real, locally quadratic near zero residual, degrade as
gracefully as \MSE{}-real (20.1--20.4~dB at $\sigma{=}50$ vs.\ 21.0;
17.2--17.3~dB on speckle~0.5 vs.\ 17.6); a small quadratic basin
suffices as mitigation. The specialization is sharpest for \MAE{}-real:
in distribution it sits with the best (leading both real columns at
seed~0, a lead within seed noise), out of distribution it is the most
brittle. For deployment (e.g., an inspection line whose sensor noise
can drift or whose incoming parts change), a range-occupancy check
(the fraction of outputs at the $[0,1]$ bounds) is a cheap
out-of-distribution sentinel, and Charbonnier or
Huber buys the graceful mode at no resolvable in-distribution cost
(Huber-real has the highest three-seed block mean, 34.9~dB). Each
family specializes to its training distribution; the two overlap only
where the distributions do (Fig.~\ref{fig:qual}). This two-way
specialization is the direct evidence that the training pair
distribution carries the inductive bias.

% =============================================================================
\section{Discussion}

\subsection{What the Data Confirms and Refutes}
\textbf{Confirmed.} (1)~H1's core: \MAE{} $>$ \MSE{} on all
synthetic columns, significant for PSNR and SSIM (Holm-corrected
Wilcoxon, Section~\ref{sec:loss-effect}), seed-replicated on 13 of 14
columns, though never exceeding 1~dB; its rider (an advantage concentrated in perceptual
metrics) is refuted, as LPIPS survives correction only under the
heaviest corruption.
(2)~H2: only the Lasso model is sparse (91\% of weights at zero) yet it
behaves like \MSE{}; sparsity and cross-noise behavior are fully
dissociated.
(3)~Poisson loss-invariance: $\le$0.7~dB spread across losses, as the
Anscombe argument predicts. (4)~H3's real-noise half, weakened:
synthetic training does not \emph{degrade} real SIDD inputs, but
forfeits $\approx$8~dB relative to matching the training distribution
($\approx$5~dB scene statistics, $\approx$3~dB noise;
Section~\ref{sec:sidd}); the recovery is loss-agnostic.

\textbf{Refuted.} H3's speckle half: no collapse occurs even at
$v{=}0.5$, where additive-trained N2N still gains $+10$~dB; our
initial framing over-predicted the brittleness of the additive
assumption. Also refuted: the tacit assumption that a robust loss buys
\emph{general} robustness. On real noise synth-trained \MAE{} is
significantly \emph{worse} than \MSE{}, and \MAE{}-real is the most
brittle model in the study out of distribution.

\subsection{Removed Confounds and Reproducibility Lessons}
\label{sec:enhancer}

An earlier draft contained four protocol defects, each of which
produced internally consistent, publishable-looking tables; only
external anchors (published baselines, the official noisy-input PSNR)
exposed them. (i)~A \texttt{PIL.ImageEnhance} post-processing stage
inflated PSNR by $+2.95$~dB and was removed as an evaluation confound.
(ii)~The training set on disk was \emph{grayscale} while evaluation was
color: the channel-statistics gap capped every learned method near
22~dB on Kodak24 regardless of loss, architecture, or step budget,
mimicking ``undertraining'', and inverted several conclusions.
(iii)~SIDD images were bicubically downsampled in training and
evaluation, averaging out part of the target noise: the noisy baseline
read 34.3~dB instead of the official 23.7~dB, turning a modest
improvement into an apparent catastrophic failure. (iv)~The BM3D
baseline was initially run per-channel (grayscale BM3D) with a
row-difference MAD $\sigma$-estimate missing its $1/\sqrt{2}$,
understating BM3D by ${\approx}2$~dB on the Gaussian columns; the
published CBM3D Kodak24 figures exposed both defects, and all BM3D
numbers here use \texttt{bm3d\_rgb} with a wavelet-domain estimate.

\subsection{Limitations}
Single consumer GPU; U-Net depth-4 backbone (absolute numbers below
specialized architectures, orderings expected stable); SIDD train/eval
slices share camera models; device-stratified splits and DND
\cite{plotz2017dnd} are future work; the full noise grid is run once per
configuration, and seed replication is partial: three seeds for
\MAE{} and \MSE{} across the whole synthetic grid
(Section~\ref{sec:loss-effect}) and per real-pair loss on the blocks
(Section~\ref{sec:sidd}), but one seed for Charbonnier, Huber, and
Lasso in the synthetic cell. An alternative we have
not tested: training
draws $\sigma_1,\sigma_2 \sim \mathcal{U}(5,50)$, so the training
residual is a Gaussian scale mixture, heavy-tailed by
construction, and part of \MAE{}'s synthetic edge may reflect robustness to this
training-set heterogeneity rather than to the test noise; a
fixed-$\sigma$ training cell would separate the two accounts.

\subsection{Reproducibility}
\label{sec:repro}
All code and scripts, the deterministic sorted-name SIDD split, and
per-image metric dumps are available at
\href{https://github.com/dyshang/noise2noise-revisited}{\nolinkurl{github.com/dyshang/noise2noise-revisited}}. Every model
trains on one consumer GPU (RTX~3080, 10~GB) in about 16 minutes; all
datasets are public, and seeds are fixed and recorded inside each
checkpoint. The seed replicates ship as per-image dumps, so
Table~\ref{tab:seed-gap} recomputes without retraining.

% =============================================================================
\section{Conclusion}

Under symmetric zero-mean training noise the population optima of \MSE{}
and \MAE{} nearly coincide (exactly so for symmetric signal posteriors),
leaving the loss-function effect in the symmetric synthetic setting
dominated by optimization
dynamics: real (sub-dB, significant, seed-replicated on 13 of 14
columns) but small, and \emph{not} mediated by weight sparsity: an explicit Lasso
penalty produces sparsity without the behavior. On the official SIDD
protocol (one sRGB smartphone benchmark), switching the training
pairs from synthetic Gaussian to the
sensor's own noisy/noisy shots, with no clean data, is worth
$\approx$8~dB for every loss ($\approx$5~dB from matched scene
statistics and $\approx$3~dB from the noise distribution,
Section~\ref{sec:sidd}), an order of magnitude more than any
in-distribution loss-function choice. The training pair distribution
sets in-distribution performance, while the loss governs
out-of-distribution robustness, where
pure \MAE{} fails catastrophically and a small quadratic basin
(Charbonnier, Huber) restores graceful degradation. On this evidence,
the training pair distribution, not the loss function, deserves the
larger share of design attention in denoising practice.
The recipe is sensor-agnostic and cheap to rerun: at roughly 16
minutes of consumer-GPU training per model
(Section~\ref{sec:repro}), each sensor or production line
can carry its own denoiser, retrained from its own paired captures
without any labeled data.
Two noisy shots per scene shift the cost from labeling to data
collection, provided the shots are pixel-registered and the signal is
static between exposures: in microscopy, photobleaching or stage drift
changes the \emph{signal}, violating the identical-signal premise of
Section~\ref{sec:theory}, as does handheld capture.
Industrial inspection meets both conditions by construction, a fixtured
part imaged by a mounted camera yielding registered repeat exposures in
seconds.
One premise must be checked per modality: fixed-pattern components
(defective pixels, nonuniformity, detector gain) are shared across
repeat exposures and are therefore retained by N2N
(Section~\ref{sec:traindist}), so the cross-shot residual-correlation
check performed there, with standard calibration upstream, is part of
the recipe.
A concrete target is multi-modal inspection of lithium-ion cells
entering second-life remanufacturing, where thermal,
short-wave-infrared, radiographic, and acoustic-imaging captures are
cheap to pair on the line but can never be obtained clean
\cite{finegan2015tomography,hsieh2015acoustic}; extending the present
protocol to these modalities, complementing reconstruction-domain
self-supervision such as Noise2Inverse
\cite{hendriksen2020noise2inverse}, is the next step of this
program.
Where paired shots are unobtainable (dynamic scenes,
video), the single-image methods of Section~\ref{sec:related} remain the recourse.

% =============================================================================

\end{document}